\documentclass[accepted]{article}

\usepackage{times}
\usepackage{graphicx} 
\usepackage{subfigure}
\usepackage{microtype}
\usepackage[scaled]{beramono}
\usepackage[T1]{fontenc}

\usepackage{natbib}

\usepackage{booktabs}
\usepackage{multirow}

\usepackage{algorithm}
\usepackage{algorithmic}

\usepackage{amsfonts}
\usepackage{amssymb}
\usepackage{paralist}

\usepackage{hyperref}

\newcommand\blfootnote[1]{%
  \begingroup
  \renewcommand\thefootnote{}\footnote{#1}%
  \addtocounter{footnote}{-1}%
  \endgroup
}

\usepackage[accepted]{icml2016}
\usepackage{xspace}
\usepackage{color}

\newcommand{\eg}{\hbox{\emph{e.g.}}\xspace}
\newcommand{\ie}{\hbox{\emph{i.e.}}\xspace}

\newcommand{\etc}{\hbox{\emph{etc.}}\xspace}
\newcommand{\vs}{\hbox{\emph{vs.}}\xspace}

\usepackage{bm}
\usepackage[normalem]{ulem}
\usepackage[version=0.96]{pgf}
\usepackage{tikz}
\usetikzlibrary{arrows,shapes,snakes,automata,backgrounds,petri,positioning,shapes.geometric,decorations.pathreplacing}

\usepackage{listings}
\newcommand\hl[2]{%
    \tikz[baseline]%
    \definecolor{col}{RGB}{255, 255, #1}%
   \node[decorate,rectangle,fill=col,anchor=text,anchor=base west,%
   outer sep=0pt,inner xsep=1pt,inner ysep=0pt, rounded corners=1pt,%
   minimum height=\ht\strutbox+1pt,draw=black!20,dashed,line width=.5pt]{
		\strut\texttt{#2}\xspace
	};%
}%
\newcommand\hlc[2]{%
    \tikz[baseline]%
    \definecolor{col}{RGB}{255, #1, 255}%
   \node[decorate,rectangle,fill=col,anchor=text,anchor=base west,%
   outer sep=0pt,inner xsep=1pt,inner ysep=0pt, rounded corners=1pt,%
   minimum height=\ht\strutbox+1pt,draw=black!20,dashed,line width=.5pt]{
		\strut\texttt{#2}\xspace
	};%
}%

\newcommand{\unk}[0]{\textsc{Unk}\xspace}

\icmltitlerunning{A Convolutional Attention Network for Extreme Summarization of Source Code}

\begin{document}

\twocolumn[
\icmltitle{A Convolutional Attention Network \\ for Extreme Summarization of Source Code}

\icmlauthor{Miltiadis Allamanis}{m.allamanis@ed.ac.uk}
\icmladdress{School of Informatics,
			University of Edinburgh, Edinburgh, EH8 9AB, United Kingdom}
\icmlauthor{Hao Peng$^\dagger$}{penghao.pku@gmail.com}
\icmladdress{School of Electronics Engineering and Computer Science,
	Peking University, Beijing 100871, China}
\icmlauthor{Charles Sutton}{csutton@inf.ed.ac.uk}
\icmladdress{School of Informatics,
			University of Edinburgh, Edinburgh, EH8 9AB, United Kingdom}
\icmlkeywords{source code naturalness, neural attention models}

\vskip 0.2in
]

\begin{abstract}
Attention mechanisms in neural networks have proved useful for problems
in which the input and output do not have fixed dimension.
Often there exist features that are locally translation invariant
and would be valuable for directing the model's attention, but previous attentional
architectures are not constructed to learn such features specifically.
We introduce an
attentional neural network that employs convolution on the input tokens to detect
local time-invariant and long-range topical attention features in a
context-dependent way. We apply this architecture to 
the problem of extreme summarization of source code snippets
into short, descriptive function name-like summaries. 
Using those features, the model sequentially generates a summary by marginalizing
over two attention mechanisms: one that
predicts the next summary token based on the attention weights of the input tokens
and another that is able to copy a code token as-is directly into the summary.
We demonstrate our convolutional attention neural network's performance on 10
popular Java projects showing that it
achieves better performance compared to previous attentional mechanisms.

\vspace{-2em}
\end{abstract}
\vspace{-2em}

\ifdefined\isaccepted
  \blfootnote{$^\dagger$Work partially done while author was as an intern at the University of Edinburgh.\vspace{-.75em}}
\fi

\section{Introduction}

Deep learning for structured prediction problems,
in which
a sequence (or more complex structure)
 of predictions need to be made given an input
 sequence, presents special difficulties,
because not only are the input and output high-dimensional, but the dimensionality is not
fixed in advance.
Recent research has tackled these problems
using neural models of attention
 \citep{mnih2014recurrent},
 which have had great recent successes in
  machine translation \citep{bahdanau2014neural} and image captioning
\citep{xu2015show}.
Attentional models have been successful because
they separate two different concerns: predicting
which input locations
are most relevant to each location
of the output; and actually predicting an output
location given the most relevant inputs.

In this paper, we suggest that
many domains contain translation-invariant features
that can help to determine the most useful
locations for attention.
For example, in a research paper, the sequence of words ``in this paper, we suggest'' often
indicates that the next few words will be
important to the topic of the paper.
As another example, suppose a neural network
is trying to predict the name of a method in the Java programming language
from its body. If we know that this method name
begins with \texttt{get} and the method body contains a statement
\texttt{return \_\_\_\_ ; }, then whatever token fills in the blank
is likely to be useful for predicting the rest of the method name.
Previous architectures for neural attention are not constructed
to learn translation-invariant features specifically.

We introduce a neural convolutional attentional model,
that includes a convolutional network
within the attention mechanism itself.
Convolutional models are a natural choice for learning translation-invariant
features  while using only a small number of parameters and for this reason
 have been highly successful in non-attentional models
for images \citep{lecun98gradient,krizhevsky2012imagenet} and
text classification \citep{blunsom2014convolutional}.
But to our knowledge they have not been applied within an attentional mechanism.
Our model uses
a set of convolutional layers --- without any pooling --- to detect patterns in the input and
identify ``interesting'' locations where attention should be focused.

We apply this network to an ``extreme'' summarization problem: We ask the network to predict
a short and descriptive name of a source code snippet (\eg a method body)
given solely its tokens.
Source
code has two distinct roles: it not only is a means of instructing a CPU to perform
a computation but also acts as an essential means
of communication among developers who need to understand, maintain and evolve
software systems.
For these reasons, software engineering research has found
that good names are important to developers
\citep{liblit2006cognitive,takang1996effects,binkley2013impact}.
Additionally, learning to summarize source code has important applications in software
engineering, such as in code understanding and in code search.
The highly structured form of source code
makes convolution naturally suited for the purpose of  extreme summarization.
Our choice of problem is inspired by previous work \citep{allamanis2015suggesting}
that tries to name existing methods (functions) using a large set of hard-coded
features, such as features from the containing class and the method signature.
But these hard-coded features may not be available for arbitrary code snippets
and in dynamically typed languages.
In contrast, in this paper we consider a more general problem: given an arbitrary
snippet of code --- without any hard-coded features --- provide
a summary, in the form of a descriptive method name.

This problem resembles a summarization task, where the method name is viewed as
the summary of the code. However, extreme
source code summarization is drastically different from natural language
summarization, because unlike natural language,
source code is unambiguous and highly structured.
Furthermore, a good summary needs
to explain \emph{how} the code instructions compose into a higher-level meaning and
not na\"{i}vely explain what the code does.
This necessitates learning higher-level
patterns in source code that uses both the structure of the code and the identifiers
to detect and explain complex code constructs.
Our extreme summarization problem may also be
viewed as a translation task, in the same way
that any summarization problem can be viewed as translation.
But a significant difference from translation is that
the input source code sequence tends to be very
large (72 on average in our data) and the output summary very small
(3 on average in our data). The length of the input sequence
necessitates the extraction of both temporally invariant attention features and
topical sentence-wide features and --- as we show in this paper ---
existing neural machine translation techniques yield sub-optimal results.

Furthermore, source code presents the challenge
of out-of-vocabulary words. Each new software project
and each new source file introduces new vocabulary about aspects
of the software's domain, data structures, and so on.
This vocabulary often does not appear in the training set.
To address this problem, we introduce a \emph{copy mechanism}, which uses the convolutional attentional
mechanism to identify important tokens in the input
even if they are out-of-vocabulary tokens that do not
appear in the training set.
The decoder, using a meta-attention mechanism, may choose to copy tokens directly from the input to the output
sequence, resembling the functionality
of \citet{vinyals2015pointer}.

The key contributions of our paper are:
\begin{inparaenum}[(a)]
\item a novel convolutional attentional
network that successfully performs extreme summarization of source code;
\item a comprehensive approach to the extreme code summarization problem, with
interest both in the machine learning and software engineering community; and
\item a comprehensive evaluation of four competing algorithms on real-world data
that demonstrates the advantage of our method compared to standard attentional
mechanisms.
\end{inparaenum}

\section{Convolutional Attention Model}
\newcommand{\attentionfeatures}[0]{\textbf{attention\_features}\xspace}
\newcommand{\attentionweights}[0]{\textbf{attention\_weights}\xspace}
\newcommand{\copyattention}[0]{\textbf{copy\_attention}\xspace}
\newcommand{\normalattention}[0]{\textbf{conv\_attention}\xspace}
\newcommand{\tab}[0]{\hspace{8pt}}

\begin{figure}[tb]
	\begin{center}
		\input{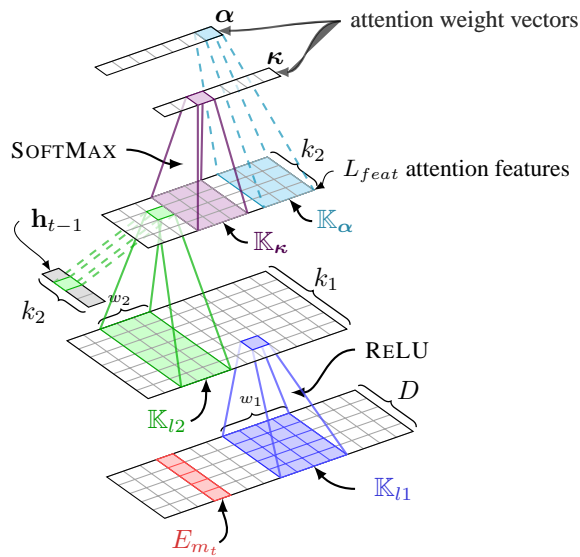}
		\caption{The architecture of the convolutional attentional network.
		\attentionfeatures learns location-specific attention
		features given an input sequence $\{m_i\}$ and a context vector $\mathbf{h}_{t-1}$.
		Given these features \attentionweights~---using
		a convolutional layer and a \textsc{SoftMax}--- computes the final attention
		weight vectors such as
		$\boldsymbol\alpha$ and $\boldsymbol\kappa$ in this figure.}
		\label{fig:architecture}
	\end{center}
\end{figure}

Our convolutional attentional model receives as input a sequence of code
subtokens\footnote{Subtokens refer to the parts of a
source code token \eg \texttt{getInputStream} has the \texttt{get}, \texttt{Input} and
\texttt{Stream} subtokens.}
$\mathbf{c}=[c_{\texttt{\tiny <S>}}, c_1, \dots c_N, c_{\tiny \texttt{</S>}}]$
and outputs an extreme summary in the form of a concise method name. The summary
is a sequence of subtokens
$\mathbf{m} = [m_{\texttt{\tiny <s>}}, m_1 \dots m_M, m_{\tiny \texttt{</s>}}]$,
where \texttt{\footnotesize <s>} and \texttt{\footnotesize </s>} are the special start and end symbols
of every subtoken sequence.
For example, in the \texttt{shouldRender} method (top left of \autoref{tbl:exampleSuggestions})
the input code subtokens are $\mathbf{c} = \left[ { \footnotesize\texttt{<S>}, \texttt{try}, \texttt{\{}, \texttt{return}, \texttt{render}, \texttt{requested}, \dots } \right]$
while the target output is $\mathbf{m} = \left[ { \footnotesize\texttt{<s>}, \texttt{should}, \texttt{render}, \texttt{</s>} }\right]$.
The neural network predicts each summary subtoken sequentially and models
$P(m_t | m_{\texttt{\tiny <s>}}, \dots, m_{t-1}, \mathbf{c})$.
Information about the previously produced subtokens $m_{\texttt{\tiny <s>}}, \dots, m_{t-1}$
is passed into a recurrent neural network that represents the input state with a vector $\mathbf{h}_{t-1}$.
Our convolutional attentional neural network (\autoref{fig:architecture}) uses
the input state $\mathbf{h}_{t-1}$ and a series of convolutions over
the embeddings of the tokens $\mathbf{c}$
to compute a matrix of attention features $L_{feat}$, (\autoref{fig:architecture})
that contains one vector of attention features for each sequence position. The resulting features are used to compute one or more normalized
attention vectors (\eg $\boldsymbol\alpha$ in \autoref{fig:architecture})
which are distributions over input token locations
containing a weight (in $\mathbb{R}^{(0,1)}$) for each subtoken in $\mathbf{c}$.
Finally, given the
weights, a context representation is computed and is used
to predict the probability distribution over the targets $m_i$. This
model is a generative bimodal model of summary text given
a code snippet.

\subsection{Learning Attention Features}
We describe our model from the bottom-up (\autoref{fig:architecture}).
First we discuss how to compute the attention features $L_{feat}$ from the
input $\mathbf{c}$ and the previous hidden state $\mathbf{h}_{t-1}$.
The basic building block of our model is a convolutional network \citep{cun1990handwritten,collobert08unified} for extracting
position and context-dependent features. The input to \attentionfeatures
is a sequence of code subtokens $\mathbf{c}$ of length
$\textsc{Len}(\mathbf{c})$ and each location is mapped to a matrix of
attention features $L_{feat}$, with size
$\left(\textsc{Len}(\mathbf{c}) + const\right) \times k_2$
where the $const$ is a fixed amount of padding.
The intuition behind \attentionfeatures is that given the input $\mathbf{c}$, it uses convolution
to compute $k_2$ features for each location.
By then using $\mathbf{h}_{t-1}$ as a multiplicative gating-like mechanism,
only the currently relevant features are kept in $L_2$.
In the final stage, we normalize $L_2$. \attentionfeatures is
described with the following pseudocode:
\begin{algorithmic}
	\STATE \attentionfeatures(code tokens $\mathbf{c}$, context $\mathbf{h}_{t-1}$)
	\STATE \tab $C \leftarrow $ \textsc{LookupAndPad}($\mathbf{c}$, $E$)
	\STATE \tab $L_1 \leftarrow$ \textsc{ReLU}(\textsc{Conv1d}($C$, $\mathbb{K}_{l1}$))
	\STATE \tab $L_2 \leftarrow \textsc{Conv1d}(L_1, \mathbb{K}_{l2}) \odot  \mathbf{h}_{t-1}$
	\STATE \tab $L_{feat} \gets L_2 / \left\Vert L_2\right\Vert_2$
	\STATE \tab \textbf{return} $L_{feat}$
\end{algorithmic}
Here $E\in\mathbb{R}^{|V| \times D}$ contains the $D$ dimensional embedding of each subtoken in names and code (\ie all possible $c_i$s and $m_i$s). The two convolution kernels are
$\mathbb{K}_{l1}\in\mathbb{R}^{D \times w_1 \times k_1}$ and
$\mathbb{K}_{l2}\in\mathbb{R}^{k_1 \times w_2 \times k_2}$,
where $w_1$, $w_2$ are the window sizes of the convolutions and \textsc{ReLU}
refers to a rectified linear unit \citep{nair2010rectified}.
The vector $\mathbf{h}_{t-1}\in\mathbb{R}^{k_2}$
represents information from the previous subtokens $m_0 \dots m_{t-1}$.
\textsc{Conv1d} performs a one-dimensional (throughout the length of sentence $\mathbf{c}$)
narrow convolution. Note that the input sequence $\mathbf{c}$ is padded
by \textsc{LookupAndPad}. The size of the padding is
such that with the narrow convolutions, the attention vector (returned by \attentionweights) has
exactly $\textsc{Len}(\mathbf{c})$ components. The $\odot$ operator is the
elementwise multiplication of a vector and a matrix, \ie $B=A\odot\mathbf{v}$
for $\mathbf{v} \in \mathbb{R}^M$ and $A$ a $M \times N$ matrix, $B_{ij}=A_{ij}v_i$.
We found the normalization of $L_2$ into $L_{feat}$ to be useful during training.
We believe it helps because of the widely varying
lengths of inputs $\mathbf{c}$. Note that no pooling happens in this model; the
input sequence $\mathbf{c}$ is of the same length as the output sequence (modulo
the padding).

To compute the final attention weight vector --- a vector with non-negative elements
and unit norm --- we define \attentionweights as a function that accepts
$L_{feat}$ from \attentionfeatures and a convolution kernel $\mathbb{K}$
of size  $k_2 \times w_3 \times 1$. \attentionweights
returns the normalized attention weights vector with length $\textsc{Len}(\mathbf{c})$
and is described by the following pseudocode:
\begin{algorithmic}
	\STATE \attentionweights(attention features $L_{feat}$, kernel $\mathbb{K}$)
	\STATE \tab \textbf{return} $\textsc{SoftMax}(\textsc{Conv1d}(L_{feat}, \mathbb{K}))$
\end{algorithmic}

\textbf{Computing the State $\mathbf{h}_t$. }
Predicting the full summary $\mathbf{m}$ is a sequential prediction problem,
where each subtoken $m_t$ is sequentially predicted given the previous state
containing information about the previous subtokens $m_0 \dots m_{t-1}$.
The state is passed through $\mathbf{h}_t \in \mathbb{R}^{k_2}$
computed by a Gated Recurrent Unit \citep{cho2014properties} \ie
\begin{algorithmic}
	\STATE GRU(current input $\mathbf{x}_t$, previous state $\mathbf{h}_{t - 1}$)
	\STATE \tab $\mathbf{r}_t \gets \sigma(\mathbf{x}_t W_{xr} + \mathbf{h}_{t - 1} W_{hr} + \mathbf{b}_r)$
	\STATE \tab $\mathbf{u}_t \gets \sigma(\mathbf{x}_t W_{xu} + \mathbf{h}_{t - 1} W_{hu} + \mathbf{b}_u)$
	\STATE \tab $\mathbf{c}_t \gets \tanh(\mathbf{x}_t W_{xc} + r_t \odot (\mathbf{h}_{t - 1} W_{hc}) + \mathbf{b}_c)$
	\STATE \tab $\mathbf{h}_t \gets (1 - \mathbf{u}_t) \odot \mathbf{h}_{t - 1} + \mathbf{u}_t \odot \mathbf{c}_t$
	\STATE \tab \textbf{return} $\mathbf{h}_t$
\end{algorithmic}
During testing the next state is computed by $\mathbf{h}_t = \textsc{GRU}(E_{m_t}, \mathbf{h}_{t - 1})$
\ie using the embedding of the current output subtoken $m_t$.
For regularization during training, we use a trick similar to \citet{bengio2015scheduled}
and with probability equal to the dropout rate we compute the next state as
$\mathbf{h}_t = \textsc{GRU}(\mathbf{\hat{n}}, \mathbf{h}_{t - 1})$, where $\mathbf{\hat{n}}$ is
the predicted embedding.

\subsection{Simple Convolutional Attentional Model}
We now use the components described above as building blocks for our extreme
summarization model. We first build \normalattention, a convolutional attentional model
that uses an attention vector $\boldsymbol\alpha$ computed from \attentionweights
to weight the embeddings of the tokens in $\mathbf{c}$ and compute the
predicted target embedding $\mathbf{\hat{n}} \in \mathbb{R}^D$.
It returns a distribution over all subtokens in $V$.
\begin{algorithmic}
	\STATE \normalattention(code $\mathbf{c}$, previous state $\mathbf{h}_{t-1}$)
        \STATE \tab $L_{feat} \leftarrow \attentionfeatures(\mathbf{c}, \mathbf{h}_{t-1})$
	\STATE \tab $\boldsymbol\alpha \leftarrow $\attentionweights$(L_{feat}, \mathbb{K}_{att})$
	\STATE \tab $\mathbf{\hat{n}} \leftarrow \sum_i \alpha_i E_{c_i}$
	\STATE \tab $\mathbf{n} \gets \textsc{SoftMax}(E\, \mathbf{\hat{n}}^{\top} + \mathbf{b})$
	\STATE \tab \textbf{return} \textsc{ToMap}$(\mathbf{n}, V)$
\end{algorithmic}
where $\mathbf{b} \in\mathbb{R}^{|V|}$ is a bias vector and \textsc{ToMap} returns a
map of each subtoken in $v_i\in V$ associated with its probability $n_i$.
We train this model using maximum likelihood. Generating from the model works as follows:
starting with the special $m_0={\footnotesize \texttt{<s>}}$ subtoken and $\mathbf{h}_0$,
at each timestep $t$ the next subtoken $m_t$ is generated using the
probability distribution $\mathbf{n}$ returned by \normalattention$(\mathbf{c}, \mathbf{h}_{t-1})$. Given the
new subtoken $m_t$, we compute the next state $\mathbf{h}_t = \textsc{GRU}(E_{m_t}, \mathbf{h}_{t - 1})$.
The process stops when the special {\footnotesize \texttt{</s>}} subtoken is generated.

\subsection{Copy Convolutional Attentional Model}
We extend \normalattention by using an additional attention vector $\boldsymbol\kappa$
as a copying mechanism that can suggest out-of-vocabulary
subtokens. In our data a significant proportion of the output
subtokens (about 35\%) appear in $\mathbf{c}$. Motivated by this,
we extend \normalattention and allow a direct copy from the
input sequence $\mathbf{c}$ into the summary. Now the network
when predicting $m_t$, with probability $\lambda$ copies a token from $\mathbf{c}$
into $m_t$ and with probability $1-\lambda$ predicts the target subtoken as in \normalattention.
Essentially, $\lambda$ acts as a \emph{meta-attention}.
When copying, a token $c_i$ is copied into $m_t$ with probability
equal to the attention weight $\kappa_i$.
The process is the following:
\begin{algorithmic}
	\STATE \copyattention(code $\mathbf{c}$, previous state $\mathbf{h}_{t-1}$)
	\STATE \tab $L_{feat} \gets$ \attentionfeatures$(\mathbf{c}, \mathbf{h}_{t-1})$
	\STATE \tab $\boldsymbol\alpha \leftarrow $\attentionweights$(L_{feat}, \mathbb{K}_{att})$
	\STATE \tab $\boldsymbol\kappa \leftarrow $\attentionweights$(L_{feat}, \mathbb{K}_{copy})$
	\STATE \tab $\lambda \leftarrow \max(\sigma(\textsc{Conv1d}(L_{feat}, \mathbb{K}_{\lambda})))$
	\STATE \tab $\mathbf{\hat{n}} \leftarrow \sum_i \alpha_i E_{c_i}$
	\STATE \tab $\mathbf{n} \gets \textsc{SoftMax}(E\, \mathbf{\hat{n}}^{\top} + \mathbf{b})$
	\STATE \tab \textbf{return} $\lambda \textsc{Pos2Voc}(\boldsymbol\kappa, \mathbf{c}) + (1 - \lambda) \textsc{ToMap}(\mathbf{n}, V)$
\end{algorithmic}
where $\sigma$ is the sigmoid function,
$\mathbb{K}_{att}$, $\mathbb{K}_{copy}$ and $\mathbb{K}_{\lambda}$ are different
convolutional kernels, $\mathbf{n} \in\mathbb{R}^{|V|}$,
$\boldsymbol\alpha, \boldsymbol\kappa\in\mathbb{R}^{\textsc{Len}(\mathbf{c})}$,
\textsc{Pos2Voc} returns a map of each subtoken in $\mathbf{c}$ (which may include out-of-vocabulary tokens)
to the probabilities $\kappa_i$ assigned by the copy mechanism.
Finally, the predictions of the two attention mechanisms are merged, returning a map that contains all potential target subtokens
in $V\cup\mathbf{c}$ and interpolating over the two attention mechanisms, using
the meta-attention weight $\lambda$.
Note that $\boldsymbol\alpha$ and $\boldsymbol\kappa$ are
analogous attention weights but are computed from different
kernels, and that $\mathbf{n}$ is computed exactly as in \normalattention.

\textbf{Objective. } To obtain
signal for the copying mechanism and $\lambda$, we input to \copyattention
a binary vector $\mathbb{I}_{\mathbf{c}=m_t}$ of size $\textsc{Len}(\mathbf{c})$
where each component is one if the code subtoken is identical to the current target subtoken $m_t$.
We can then compute the probability of a correct copy over the marginalization
of the two mechanisms, \ie
$$P(m_t|\mathbf{h}_{t-1}, \mathbf{c})
  = \lambda \sum_i \kappa_i \mathbb{I}_{c_i=m_t} + (1-\lambda)\mu r_{m_t}$$
where the first term is the probability of a correct copy (weighted by $\lambda$)
and the second term is the probability of the target token $m_t$ (weighted by $1-\lambda$).
We use $\mu \in (0, 1]$ to penalize the model when the simple attention predicts
an \unk but the subtoken can be predicted exactly by the copy mechanism,
otherwise $\mu=1$. We arbitrarily used $\mu=e^{-10}$, although variations did not
affect performance.

\subsection{Predicting Names}
To predict a full method name, we use a hybrid breath-first search and beam search. We
start from the special $m_0={\footnotesize\texttt{<s>}}$ subtoken and maintain a (max-)heap
of the highest probability partial predictions so far. At each step, we pick
the highest probability prediction and predict its next subtokens, pushing
them back to the heap. When the {\footnotesize\texttt{</s>}} subtoken is generated the suggestion
is moved onto the list of suggestions. Since we are interested in the
top $k$ suggestions, at each point, we prune partial suggestions
that have a probability less than the current best $k$th full suggestion. To make the
process tractable, we limit the partial suggestion heap size and stop
iterating after 100 steps.

\section{Evaluation}
\textbf{Dataset Collection. } We are interested in the extreme summarization
problem where we summarize a source code snippet into a short and concise
method-like name. Although such a dataset does not exist for arbitrary snippets
of source code, it is natural to consider existing method (function) bodies as
our snippets and the method names picked by the developers as our target extreme
summaries.

To collect a good dataset of good quality, we
cloned 11 open source Java projects from \href{http://github.com}{GitHub}. We
obtained the most popular projects by taking the sum of the $z$-scores of
the number of watchers and forks of each project, using GHTorrent \citep{gousios2012ghtorrent}.
We selected the top 11 projects that
contained more than 10MB of source code files each and use
\textsf{libgdx} as a development set. These projects have thousands of forks
and stars, being widely known among software developers.
The projects along with short descriptions
are shown in \autoref{tbl:dataset}. We used this procedure to
select a mature, large, and diverse corpus of
real source code. For each file, we extract the Java methods, removing
methods that are overridden, are abstract or are the constructors of a class.
We find the overridden methods by an
approximate static analysis that checks for inheritance relationships
and the \texttt{@Override} annotation. Overridden methods are removed,
since they are highly repetitive and their names are easy to predict. Any full tokens that are
identical to the method name (\eg in recursion) are replaced with a special
\textsc{Self} token. We split and lowercase each method name and code token into
subtokens $\{m_i\}$ and $\{c_i\}$ on \texttt{camelCase} and
\texttt{snake\_case}. The dataset and code can be found at \href{http://groups.inf.ed.ac.uk/cup/codeattention/}{groups.inf.ed.ac.uk/cup/codeattention}.

\begin{table*}[tb]
	\begin{center}
		\scriptsize
		\begin{tabular}{lrp{3.1	cm}rrrrrrrr} \toprule
			\multirow{3}{*}{Project Name} & \multirow{3}{*}{Git SHA} & \multirow{3}{*}{Description} & \multicolumn{8}{c}{F1}\\
			&&& \multicolumn{2}{c}{tf-idf} & \multicolumn{2}{c}{Standard Attention} & \multicolumn{2}{c}{\normalattention} & \multicolumn{2}{c}{\copyattention} \\
			&&& Rank 1 & Rank 5 & Rank 1 & Rank 5& Rank 1 & Rank 5& Rank 1 & Rank 5 \\ \cmidrule(r){1-3} \cmidrule(r){4-5} \cmidrule(r){6-7} \cmidrule(r){8-9} \cmidrule(r){10-11}
			\textsf{cassandra} & \texttt{53e370f} & Distributed Database
						& 40.9 & 52.0 & 35.1 & 45.0 & 46.5 & 60.0 & \textbf{48.1} & \textbf{63.1}\\
			\textsf{elasticsearch} & \texttt{485915b} & REST Search Engine
						& 27.8 & 39.5 & 20.3 & 29.0 & 30.8 & 45.0 & \textbf{31.7} & \textbf{47.2}\\
			\textsf{gradle} & \texttt{8263603} & Build System
						& 30.7 & 45.4 & 23.1 & 37.0 & 35.3 & 52.5 & \textbf{36.3} & \textbf{54.0}\\
			\textsf{hadoop-common} & \texttt{42a61a4} & Map-Reduce Framework
						& 34.7 & 48.4 & 27.0 & 45.7 & 38.0 & 54.0 & \textbf{38.4} & \textbf{55.8}\\
			\textsf{hibernate-orm} & \texttt{e65a883} & Object/Relational Mapping
						& 53.9 & 63.6 & 49.3 & 55.8 & 57.5 & 67.3 & \textbf{58.7} & \textbf{69.3}\\
			\textsf{intellij-community} & \texttt{d36c0c1} & IDE
						& 28.5 & 42.1 & 23.8 & 41.1 & 33.1 & 49.6 & \textbf{33.8} & \textbf{51.5}\\
			\textsf{liferay-portal} & \texttt{39037ca} & Portal Framework
						& 59.6 & 70.8 & 55.4 & 70.6 & 63.4 & 75.5 & \textbf{65.9} & \textbf{78.0}\\
			\textsf{presto} & \texttt{4311896} & Distributed SQL query engine
						& 41.8 & 53.2 & 33.4 & 41.4 & 46.3 & 59.0 & \textbf{46.7} & \textbf{60.2}\\
			\textsf{spring-framework} & \texttt{826a00a} & Application Framework
						& 35.7 & 47.6 & 29.7 & 41.3 & 35.9 & 49.7 & \textbf{36.8} & \textbf{51.9}\\
			\textsf{wildfly} & \texttt{c324eaa} & Application Server
						& 45.2 & 57.7 & 32.6 & 44.4 & \textbf{45.5} & 61.0 & 44.7 & \textbf{61.7}\\
			\bottomrule
		\end{tabular}
		\caption{Open source Java projects used and F1 scores achieved. Standard attention
		refers to the model of \citet{bahdanau2014neural}.}\label{tbl:dataset}
	\end{center}
\end{table*}

\textbf{Experimental Setup. }
To measure the quality of our suggestions we compute two scores.
\emph{Exact match} is the percentage of the method names
predicted exactly, while the F1 score is computed
in a per-subtoken basis.
When suggesting summaries, each model returns
a ranked list. We compute exact match and F1 at rank 1 and
5, as the best score achieved by any one of the top suggestions (\ie if the fifth
suggestion achieves the best F1 score, we use this one for computing
F1 at rank 5).
Using BLEU \citep{papineni02bleu} would have been possible,
but it would not be different from F1 given the short lengths of our output sequences (3 on average).
We use each project separately, training one network for
each project and testing on the respective test set. This is because
each project's domain varies widely and little information can
be transferred among them, due to the principle of code reusability
of software engineering.  We note that we attempted
to train a single model using all project training sets
but this yielded significantly worse results for all algorithms.
For each project, we split the \emph{files} (top-level Java classes) uniformly at random into
training (65\%), validation (5\%) and test (30\%) sets.
We optimize hyperparameters using Bayesian optimization with
Spearmint \citep{snoek2012practical} maximizing F1 at rank 5.

For comparison, we use two algorithms: a tf-idf algorithm
that computes a tf-idf vector from the code snippet subtokens and
suggests the names of the nearest neighbors using cosine similarity.
We also use the standard attention model of \citet{bahdanau2014neural}
that uses a biRNN and fully connected components, that has been successfully used
in machine translation. 
We perform hyperparameter optimizations following the same protocol on \textsf{libgdx}.

\textbf{Training. }
To train \normalattention and \copyattention we optimize the objective using stochastic
gradient descent with RMSProp and Nesterov momentum \citep{sutskever2013importance,hinton2012neural}.
We use dropout \citep{srivastava2014dropout} on all parameters, parametric leaky \textsc{ReLU}s
\citep{maas2013rectified, he2015delving} and gradient clipping.
Each of the parameters of the model is initialized with normal random noise
around zero, except for $\mathbf{b}$ that is initialized to the log of
the empirical frequency of each target token in the training set.
For \normalattention the optimized hyperparameters are $k_1=k_2=8$,
$w_1=24$, $w_2=29$, $w_3=10$, dropout rate 50\% and $D=128$. For \copyattention
the optimized hyperparameters are $k_1=32$, $k_2=16$, $w_1=18$, $w_2=19$,
$w_3=2$, dropout rate 40\% and $D=128$.

\subsection{Quantitative Evaluation}

\begin{table*}[tb]
	\begin{center}
		\footnotesize
		\begin{tabular}{lrrrrrrrrrrrr} \toprule
		& \multicolumn{2}{c}{F1 (\%)} && \multicolumn{2}{c}{Exact Match (\%)}
					&& \multicolumn{2}{c}{Precision (\%)} && \multicolumn{2}{c}{Recall (\%)} \\
		& Rank 1 & Rank 5 && Rank 1 & Rank 5 	&& Rank 1 & Rank 5 && Rank 1 & Rank 5	\\ \cmidrule{2-3} \cmidrule{5-6} \cmidrule{8-9} \cmidrule{11-12}
		tf-idf & 40.0 & 52.1 && \textbf{24.3} & 29.3 && 41.6 & 55.2 && \textbf{41.8} & 51.9\\
		Standard Attention& 33.6 & 45.2 && 17.4 & 24.9 && 35.2 & 47.1 && 35.1 & 42.1\\
		\normalattention& 43.6 & 57.7 && 20.6 & 29.8 && 57.4 & 73.7 && 39.4 & 51.9\\
		\copyattention& \textbf{44.7} & \textbf{59.6} && 23.5 & \textbf{33.7} && \textbf{58.9} & \textbf{74.9} && 40.1 & \textbf{54.2}\\
		 \bottomrule
	 \end{tabular}
		\caption{Evaluation metrics averaged across projects. Standard Attention
		refers to the work of \citet{bahdanau2014neural}.}\label{tbl:evaluation}
	\end{center}
\end{table*}

\autoref{tbl:dataset} shows the F1 scores achieved by the different methods
for each project while \autoref{tbl:evaluation} shows a quantitative evaluation, averaged
across all projects. ``Standard Attention'' refers to the machine translation model of
\citet{bahdanau2014neural}. The tf-idf algorithm seems to be performing very
well, showing that the bag-of-words representation of the input code is a strong
indicator of its name. Interestingly, the standard attention model
performs worse than tf-idf in this domain,
while \normalattention and \copyattention perform the best. The copy mechanism
gives a good F1 improvement at rank 1 and a larger improvement at rank 5. Although
our convolutional attentional models have an exact match similar to tf-idf, they
achieve a much higher precision compared to all other algorithms.

These differences in the data characteristics
could be the cause of the low performance achieved by the model of
\citet{bahdanau2014neural}. Although source code
snippets resemble natural language sentences,
they are more structured, much longer and vary greatly in length.
In our training sets, each method has on average 72 tokens (median 25 tokens,
standard deviation 156) and the output method names are made up from 3
subtokens on average ($\sigma=1.7$).

\textbf{\textsc{OoV} Accuracy. }
We measure the out-of-vocabulary (\textsc{OoV}) word accuracy as the percentage of the out-of-vocabulary subtokens
that are correctly predicted by \copyattention. On average, across our dataset,
4.4\% of the test method name subtokens are \textsc{OoV}.
Naturally, the standard attention model and tf-idf have an \textsc{OoV}
accuracy of zero, since they are unable to predict those tokens.
On average we get a 10.5\% \textsc{OoV} accuracy at rank 1 and 19.4\% at rank 5.
This shows that the copying mechanism is useful in this domain and especially
in smaller projects that tend to have more \textsc{OoV} tokens. We also note
that \textsc{OoV} accuracy varies across projects, presumably due to
different coding styles.

\textbf{Topical \vs Time-Invariant Feature Detection. }
The difference of the performance between the \copyattention and
the standard attention model of \citet{bahdanau2014neural} raises an interesting
question. What does \copyattention learn that cannot be learned by the standard
attention model? One hypothesis is
that the biRNN of the standard attention model fails to capture long-range
features, especially in very long inputs. To test our hypothesis, we shuffle the subtokens in
\textsf{libgdx}, essentially removing all features that depend on the sequential information.
Without any local features all models should reduce to achieving performance similar to tf-idf.
Indeed, \copyattention now has an F1 at rank 1 that is +1\% compared to
tf-idf (presumably thanks to the language model-like structure
of the output), while the standard attention model worsens its performance
getting an F1 score (rank 1) of 26.2\%, compared to the original 41.8\%. This
suggests that the biRNN fails to capture long-range topical attention features.

\textbf{A simpler $\mathbf{h}_{t-1}$. }
Since the target summaries are
quite short, we tested a simpler alternative to the GRU, assigning
$\mathbf{h}_{t-1}=W \times [G_{m_{t-1}}, G_{m_{t-2}}]$,
where $G \in \mathbb{R}^{D \times |V|}$ is a new embedding matrix (different
from the embeddings in $E$) and $W$ is a $k_2 \times D \times 2$ tensor. This model
is simpler and slightly faster to train and achieves similar performance to \copyattention,
reducing F1 by less than 1\%.

\subsection{Qualitative Evaluation}

\begin{figure*}[tb]
\begin{center}
\begin{tabular}{llrlr} \toprule
	\multicolumn{2}{c}{Target} & \multicolumn{2}{c}{Attention Vectors} & $\lambda$\\ \midrule
	\multirow{2}{*}{$m_1$} & \multirow{2}{*}{\texttt{set}} & $\boldsymbol\alpha=$& {\footnotesize \hl{227}{<s>}\hl{216}{\{} \hl{210}{this}\hl{198}{.}\hl{209}{use}\hl{135}{\uline{Browser}}\hl{0}{Cache} \hl{142}{=} \hl{230}{use}\hl{241}{\uline{Browser}}\hl{252}{Cache}\hl{254}{;} \hl{254}{\}}\hl{253}{</s>} } & \multirow{2}{*}{0.012}\\
	 && $\boldsymbol\kappa=$& {\footnotesize \hlc{249}{<s>}\hlc{249}{\{} \hlc{247}{this}\hlc{231}{.}\hlc{190}{use}\hlc{227}{\uline{Browser}}\hlc{213}{Cache} \hlc{210}{=} \hlc{230}{use}\hlc{247}{\uline{Browser}}\hlc{252}{Cache}\hlc{254}{;} \hlc{254}{\}}\hlc{254}{</s>}}\\ \midrule

   \multirow{2}{*}{$m_2$} & \multirow{2}{*}{\texttt{use}} & $\boldsymbol\alpha=$& {\footnotesize \hl{242}{<s>}\hl{237}{\{} \hl{242}{this}\hl{0}{.}\hl{219}{use}\hl{239}{\uline{Browser}}\hl{245}{Cache} \hl{167}{=} \hl{200}{use}\hl{235}{\uline{Browser}}\hl{241}{Cache}\hl{238}{;} \hl{225}{\}}\hl{230}{</s>} } & \multirow{2}{*}{0.974}\\
	 && $\boldsymbol\kappa=$& {\footnotesize \hlc{254}{<s>}\hlc{254}{\{} \hlc{254}{this}\hlc{254}{.}\hlc{2}{use}\hlc{253}{\uline{Browser}}\hlc{254}{Cache} \hlc{254}{=} \hlc{253}{use}\hlc{254}{\uline{Browser}}\hlc{254}{Cache}\hlc{254}{;} \hlc{254}{\}}\hlc{254}{</s>}}\\ \midrule

   \multirow{2}{*}{$m_3$} &  \multirow{2}{*}{\texttt{browser}} & $\boldsymbol\alpha=$& {\footnotesize \hl{199}{<s>}\hl{170}{\{} \hl{197}{this}\hl{0}{.}\hl{215}{use}\hl{230}{\uline{Browser}}\hl{242}{Cache} \hl{162}{=} \hl{171}{use}\hl{220}{\uline{Browser}}\hl{215}{Cache}\hl{148}{;} \hl{55}{\}}\hl{103}{</s>} } & \multirow{2}{*}{0.969}\\
    && $\boldsymbol\kappa=$& {\footnotesize \hlc{254}{<s>}\hlc{254}{\{} \hlc{254}{this}\hlc{254}{.}\hlc{250}{use}\hlc{23}{\uline{Browser}}\hlc{242}{Cache} \hlc{254}{=} \hlc{254}{use}\hlc{250}{\uline{Browser}}\hlc{254}{Cache}\hlc{254}{;} \hlc{254}{\}}\hlc{254}{</s>}}\\ \midrule

   \multirow{2}{*}{$m_4$} & \multirow{2}{*}{\texttt{cache}} & $\boldsymbol\alpha=$& {\footnotesize \hl{213}{<s>}\hl{177}{\{} \hl{193}{this}\hl{132}{.}\hl{213}{use}\hl{240}{\uline{Browser}}\hl{249}{Cache} \hl{207}{=} \hl{182}{use}\hl{231}{\uline{Browser}}\hl{231}{Cache}\hl{106}{;} \hl{0}{\}}\hl{100}{</s>} } & \multirow{2}{*}{0.583}\\
	 && $\boldsymbol\kappa=$& {\footnotesize \hlc{254}{<s>}\hlc{254}{\{} \hlc{254}{this}\hlc{254}{.}\hlc{254}{use}\hlc{203}{\uline{Browser}}\hlc{63}{Cache} \hlc{254}{=} \hlc{254}{use}\hlc{252}{\uline{Browser}}\hlc{247}{Cache}\hlc{254}{;} \hlc{254}{\}}\hlc{254}{</s>}}\\ \midrule

   \multirow{2}{*}{$m_5$} & \multirow{2}{*}{\textsc{End}} & $\boldsymbol\alpha=$& {\footnotesize \hl{215}{<s>}\hl{181}{\{} \hl{190}{this}\hl{175}{.}\hl{203}{use}\hl{239}{\uline{Browser}}\hl{249}{Cache} \hl{218}{=} \hl{185}{use}\hl{230}{\uline{Browser}}\hl{230}{Cache}\hl{86}{;} \hl{0}{\}}\hl{102}{</s>} } & \multirow{2}{*}{0.066}\\
	 && $\boldsymbol\kappa=$& {\footnotesize \hlc{254}{<s>}\hlc{254}{\{} \hlc{254}{this}\hlc{254}{.}\hlc{254}{use}\hlc{243}{\uline{Browser}}\hlc{30}{Cache} \hlc{250}{=} \hlc{254}{use}\hlc{252}{\uline{Browser}}\hlc{243}{Cache}\hlc{254}{;} \hlc{254}{\}}\hlc{254}{</s>}}\\ \bottomrule

\end{tabular}
\end{center}

\caption{Visualization of \copyattention used to compute $P(m_t|m_{0} \dots m_{t-1}, \mathbf{c})$
for \texttt{setUseBrowserCache} in \textsf{libgdx}. The
darker the color of a subtoken, they higher its attention weight.
This relationship is linear.
Yellow indicates the convolutional attention weight of the \normalattention component, while purple the attention
of the copy mechanism. Since the values of $\boldsymbol\alpha$ are usually spread across the tokens
the colors show a normalized $\boldsymbol\alpha$, \ie $\boldsymbol\alpha / \left\Vert\boldsymbol\alpha\right\Vert_\infty$.
In contrast, the copy attention weights $\boldsymbol\kappa$ are usually very
peaky and we plot them as-is. Underlined subtokens are out-of-vocabulary. $\lambda$ shows
the meta-attention probability of using the copy attention $\boldsymbol\kappa$ \vs the
convolutional attention $\boldsymbol\alpha$. More visualizations of \textsf{libgdx}
methods can be found at \url{http://groups.inf.ed.ac.uk/cup/codeattention/}.}
\label{fig:attentionsample}
\end{figure*}

\begin{figure*}[tb]
	\begin{center}
\begin{tabular}{llrlr} \toprule
	\multicolumn{2}{c}{Target} & \multicolumn{2}{c}{Attention Vectors} & $\lambda$\\ \midrule
	\multirow{2}{*}{$m_1$} & \multirow{2}{*}{\texttt{is}}
	  & $\boldsymbol\alpha=$& {\footnotesize \hl{194}{<s>}\hl{171}{\{} \hl{161}{return} \hl{168}{(}\hl{149}{m}\hl{62}{Flags} \hl{72}{\&} \hl{92}{e}\hl{137}{Bullet}\hl{197}{Flag}\hl{146}{)} \hl{0}{==} \hl{48}{e}\hl{173}{Bullet}\hl{249}{Flag}\hl{254}{;} \hl{254}{\}}\hl{253}{</s>} } & \multirow{2}{*}{0.012}\\
	 && $\boldsymbol\kappa=$& {\footnotesize \hlc{251}{<s>}\hlc{248}{\{} \hlc{248}{return} \hlc{247}{(}\hlc{245}{m}\hlc{226}{Flags} \hlc{221}{\&} \hlc{236}{e}\hlc{229}{Bullet}\hlc{244}{Flag}\hlc{247}{)} \hlc{233}{==} \hlc{212}{e}\hlc{225}{Bullet}\hlc{251}{Flag}\hlc{254}{;} \hlc{254}{\}}\hlc{254}{</s>}}\\ \midrule

   \multirow{2}{*}{$m_2$} & \multirow{2}{*}{\texttt{bullet}}
	  & $\boldsymbol\alpha=$& {\footnotesize \hl{211}{<s>}\hl{220}{\{} \hl{209}{return} \hl{217}{(}\hl{135}{m}\hl{89}{Flags} \hl{185}{\&} \hl{129}{e}\hl{59}{Bullet}\hl{191}{Flag}\hl{205}{)} \hl{76}{==} \hl{0}{e}\hl{133}{Bullet}\hl{212}{Flag}\hl{218}{;} \hl{168}{\}}\hl{177}{</s>} } & \multirow{2}{*}{0.436}\\
	 && $\boldsymbol\kappa=$& {\footnotesize \hlc{254}{<s>}\hlc{254}{\{} \hlc{254}{return} \hlc{254}{(}\hlc{254}{m}\hlc{243}{Flags} \hlc{249}{\&} \hlc{254}{e}\hlc{249}{Bullet}\hlc{252}{Flag}\hlc{254}{)} \hlc{254}{==} \hlc{235}{e}\hlc{52}{Bullet}\hlc{246}{Flag}\hlc{254}{;} \hlc{254}{\}}\hlc{254}{</s>}}\\ \midrule

  \multirow{2}{*}{$m_3$} & \multirow{2}{*}{\textsc{End}}
	  & $\boldsymbol\alpha=$& {\footnotesize \hl{202}{<s>}\hl{215}{\{} \hl{208}{return} \hl{214}{(}\hl{186}{m}\hl{189}{Flags} \hl{220}{\&} \hl{197}{e}\hl{168}{Bullet}\hl{174}{Flag}\hl{209}{)} \hl{177}{==} \hl{180}{e}\hl{225}{Bullet}\hl{232}{Flag}\hl{172}{;} \hl{0}{\}}\hl{118}{</s>} } & \multirow{2}{*}{0.174}\\
	 && $\boldsymbol\kappa=$& {\footnotesize \hlc{254}{<s>}\hlc{254}{\{} \hlc{254}{return} \hlc{254}{(}\hlc{254}{m}\hlc{254}{Flags} \hlc{253}{\&} \hlc{254}{e}\hlc{253}{Bullet}\hlc{253}{Flag}\hlc{254}{)} \hlc{254}{==} \hlc{254}{e}\hlc{189}{Bullet}\hlc{71}{Flag}\hlc{254}{;} \hlc{254}{\}}\hlc{254}{</s>}}\\ \midrule

\end{tabular}
\end{center}

	\caption{Visualization of \copyattention modeling
	$P(m_t|m_{0} \dots m_{t-1}, \mathbf{c})$ for \texttt{isBullet} in \textsf{libgdx}.
	The \copyattention captures location-invariant
	features and the topicality of the input code sequence.
	For information about the visualization see \autoref{fig:attentionsample}.}
	\label{fig:attentionsample2}
\end{figure*}

\begin{table*}[tbp]
\begin{center}
	\newcommand{\methodsig}[2]{\lstinline[basicstyle=\footnotesize\ttfamily\bfseries]+#2+}
\newcommand{\suggest}[2]{ {\footnotesize$\tiny\color{black!30}\blacktriangleright$\texttt{#1}~(#2\%)} }

\begin{tabular}{p{0.94\columnwidth}p{1.06\columnwidth}} \toprule
\methodsig{a}{boolean shouldRender()} & \methodsig{b}{void reverseRange(Object[] a, int lo, int hi)} \\ \cmidrule(r){1-1} \cmidrule(r){2-2}
\vspace{-1em}\begin{lstlisting}
try {
  return renderRequested||isContinuous;
} finally {
  renderRequested = false;
}
\end{lstlisting} \vspace{-1em}&
 \vspace{-1em}\begin{lstlisting}
hi--;
while (lo < hi) {
  Object t = a[lo];
  a[lo++] = a[hi];
  a[hi--] = t;
}
\end{lstlisting} \vspace{-1em} \\
\hangindent=0.3cm\uline{Suggestions}: \suggest{is,render}{27.3}  \suggest{is,continuous}{10.6}
\suggest{is,requested}{8.2}  \suggest{render,continuous}{6.9}
\suggest{get,render}{5.7}\vspace{.75em}
&
\hangindent=0.3cm\uline{Suggestions}: \suggest{reverse,range}{22.2}  \suggest{reverse}{13.0}
\suggest{reverse,lo}{4.1}  \suggest{reverse,hi}{3.2}  \suggest{merge,range}{2.0}\vspace{.75em}
\\  \toprule
\methodsig{c}{int createProgram()} & \methodsig{d}{VerticalGroup right()}\\ \cmidrule(r){1-1} \cmidrule(r){2-2}
\vspace{-1em}\begin{lstlisting}
GL20 gl = Gdx.gl20;
int program = gl.glCreateProgram();
return program != 0 ? program : -1;
\end{lstlisting}\vspace{-1em}
&
\vspace{-1em}\begin{lstlisting}
align |= Align.right;
align &= ~Align.left;
return this;
\end{lstlisting}\vspace{-1em} \\
\hangindent=0.3cm\uline{Suggestions}: \suggest{create}{18.36}  \suggest{init}{7.9}
\suggest{render}{5.0}  \suggest{initiate}{5.0}  \suggest{load}{3.4}\vspace{.5em}
&
\hangindent=0.3cm\uline{Suggestions}: \suggest{left}{21.8}  \suggest{top}{21.1}
\suggest{right}{19.5}  \suggest{bottom}{18.5}  \suggest{align}{3.7}\vspace{.5em}
\\  \toprule
\methodsig{e}{boolean isBullet()} & \methodsig{f}{float getAspectRatio()}\\ \cmidrule(r){1-1} \cmidrule(r){2-2}
\vspace{-1em}\begin{lstlisting}
return (m_flags & e_bulletFlag)
         == e_bulletFlag;
\end{lstlisting} \vspace{-1em}
&
\vspace{-1em}\begin{lstlisting}
return (height == 0) ?
     Float.NaN : width / height;
\end{lstlisting}\vspace{-1em} \\
\hangindent=0.3cm\uline{Suggestions}: \suggest{is}{13.5}  \suggest{is,bullet}{5.5}
\suggest{is,enable}{5.1}  \suggest{enable}{2.8}  \suggest{mouse}{2.7}\vspace{.75em}
&
\hangindent=0.3cm\uline{Suggestions}: \suggest{get,{\sffamily\textsc{Unk}}}{9.0}  \suggest{get,height}{8.7}
\suggest{get,width}{6.5}  \suggest{get}{5.7}  \suggest{get,size}{4.2}\vspace{.75em}
\\  \toprule
\methodsig{g}{int minRunLength(int n)} & \methodsig{h}{JsonWriter pop()}\\ \cmidrule(r){1-1} \cmidrule(r){2-2}
\vspace{-1em}\begin{lstlisting}
if (DEBUG) assert n >= 0;
int r = 0;
while (n >= MIN_MERGE) {
  r |= (n & 1);
  n >>= 1;
}
return n + r;
\end{lstlisting} \vspace{-1em}
&
\vspace{-1em}\begin{lstlisting}
if (named) throw
    new IllegalStateException((*@\emph{\textsf{\textsc{UnkString}}}@*));
stack.pop().close();
current = stack.size == 0 ?
             null : stack.peek();
return this;
\end{lstlisting}\vspace{-1em} \\
\hangindent=0.3cm\uline{Suggestions}: \suggest{min}{43.7}  \suggest{merge}{13.0}
\suggest{pref}{1.9}  \suggest{space}{1.0}  \suggest{min,all}{0.8}
&
\hangindent=0.3cm\uline{Suggestions}: \suggest{close}{21.4}  \suggest{pop}{10.2}
\suggest{first}{6.5}  \suggest{state}{3.8}  \suggest{remove}{2.2} \\ \toprule
\methodsig{i}{Rectangle setPosition(float x, float y)} & \methodsig{j}{float surfaceArea()} \\ \cmidrule(r){1-1} \cmidrule(r){2-2}
\vspace{-1em}\begin{lstlisting}
this.x = x;
this.y = y;
return this;
\end{lstlisting}\vspace{-1em}
&
\vspace{-1em}\begin{lstlisting}
return 4 * MathUtils.PI *
       this.radius * this.radius;
\end{lstlisting}\vspace{-1em}
\\
\hangindent=0.3cm\uline{Suggestions}: \suggest{set}{54.0}  \suggest{set,y}{12.8}  \suggest{set,x}{9.0}
\suggest{set,position}{8.6}  \suggest{set,bounds}{1.68}\vspace{.5em}
&
\hangindent=0.3cm\uline{Suggestions}: \suggest{dot,radius}{26.5}  \suggest{dot}{13.1}
\suggest{crs,radius}{9.0}  \suggest{dot,circle}{6.5}  \suggest{crs}{4.1}\vspace{.5em}
\\ \bottomrule
\end{tabular}

	\caption{A sample of handpicked snippets (the sample is necessarily
	 limited to short methods because of space limitations)
	and the respective suggestions that illustrate some
	interesting challenges of the domain and how the \copyattention model
	handles them or fails. Note that the algorithms
	do \emph{not} have access to the signature of the method but only
	to the body. Examples taken from
	the \textsf{libgdx} Android/Java graphics library test set.}
	\label{tbl:exampleSuggestions}
\end{center}
\end{table*}

\autoref{fig:attentionsample} shows a visualization of a
small method that illustrates how \copyattention typically works.
At the first step,
it focuses its attention at the whole method and decides upon the
first subtoken. In a large number of cases this includes subtokens such as \texttt{get},
\texttt{set}, \texttt{is}, \texttt{create} \etc In the next steps the
meta-attention mechanism is highly confident about the copying mechanism ($\lambda=0.97$ in \autoref{fig:attentionsample}) and
sequentially copies the correct subtokens from the code snippet into the name.
We note that across many examples the copying mechanism tends to have
a significantly more focused attention vector $\boldsymbol\kappa$,
compared to the attention vector $\boldsymbol\alpha$. Presumably, this
happens because of the different training
signals of the attention mechanisms.

A second example of \copyattention is seen in \autoref{fig:attentionsample2}.
Although due to space limitations this is a relatively short method, it illustrates
how the model has learned both time-invariant features and topical features.
It correctly detects the
\lstinline+==+ operator and predicts that the method has a high
probability of starting with \texttt{is}. Furthermore, in the next step
(prediction of the $m_2$ \texttt{bullets} subtoken) it successfully learns to
ignore the \texttt{e} prefix (prepended on all enumeration variables in that project)
and the \texttt{flag} subtoken that does not provide useful information for the
summary.

\autoref{tbl:exampleSuggestions} presents a set of hand-picked
examples from \textsf{libgdx} that show interesting challenges
of the domain and how our \copyattention handles them.
Understandably, the model does \emph{not} distinguish between \texttt{should} and \texttt{is}
--- both implying a \lstinline+boolean+ return value --- and instead of
\texttt{shouldRender}, \texttt{isRender} is suggested.
The \texttt{getAspectRatio}, \texttt{surfaceArea} and \texttt{minRunLength} examples show the
challenges of describing a previously unseen abstraction. Interestingly,
the model correctly recognizes that a novel (\unk) token should be predicted after \texttt{get}
in \texttt{getAspectRatio}. Most surprisingly, \texttt{reverseRange} is predicted
correctly, because of the structure of the code,
even though no code tokens contain the summary subtokens.

\section{Related Work}
Convolutional neural networks have been used for image classification
with great success \citep{krizhevsky2012imagenet,szegedy2015going,cun1990handwritten,lecun98gradient}.
More related to this work is the use of convolutional neural networks for text
classification \citep{blunsom2014convolutional}. Closely related is the
work of \citet{denil2014modelling} that learns representations of documents
using convolution but uses the network activations to summarize a document rather
than an attentional model.
\citet{rush2015neural} use an attention-based encoder to summarize sentences,
but do not use convolution for their attention mechanism.
Our work is also related to other work in attention mechanisms for
text \citep{hermann2015teaching} and images \citep{xu2015show,mnih2014recurrent}
that does not use convolution to provide the attention values.
Pointer networks \citep{vinyals2015pointer} are similar to our copy mechanism
but use an RNN for providing attention.
Finally, distantly related to this work is research on neural
architectures that learn code-like behaviors
\citep{graves2014neural,zaremba2014learning,joulin2015inferring,grefenstette2015learning,dyer2015transition,reed2015neural,neelakantan2015neural}.

In recent years, thanks to the insight of \citet{hindle2012naturalness}
the use of probabilistic models for software engineering applications
has grown. Research has mostly focused on token-level \citep{nguyen2013statistical,tu2014localness}
and syntax-level \citep{maddison2014structured} language models of code
and translation between programming languages~\citep{karaivanov2014phrase,nguyen2014migrating}.
\citet{movshovitz2013natural} learns to predict code comments using
a source code topic model. \citet{allamanis2015bimodal} create a generative model of
source code given a natural language query and \citet{oda2015learning}
use machine translation to convert source code into pseudocode.
Closer to our work, \citet{raychev2015predicting} aim to predict
names and types of variables, whereas \citet{allamanis2014learning} and
\citet{allamanis2015suggesting} suggest names for variables, methods and
classes. Similar to \citet{allamanis2015suggesting}, we predict
method names but using only the tokens within a method and no other
features (\eg method signature). \citet{mou2016convolutional} use syntax-level
convolutional neural networks to learn vector
representations for code and classify student submissions into
tasks without considering naming. \citet{piech2015learning} also learn program embeddings from
student submissions using the program state, to assist MOOC students debug
their submissions but do not consider naming. Additionally, compared to
\citet{piech2015learning} and \citet{mou2016convolutional} our work looks into
highly diverse, non-student submission code
that performs a wide range of real-world tasks.

\section{Discussion \& Conclusions}

Modeling and understanding source code
artifacts through machine learning can have a direct impact in
software engineering research. The problem of extreme code summarization is a first
step towards the more general goal of developing machine learning representations
of source code that will allow machine learning methods to reason probabilistically
about code resulting in useful software engineering tools that will help
code construction and maintenance.

Additionally, source code --- and its derivative artifacts --- represent a new modality for
machine learning with very different characteristics compared
to images and natural language. Therefore, models of source code necessitate
research into new methods that could have interesting parallels to images and natural language.
This work is a step towards this direction: our neural convolutional
attentional model attempts to ``understand'' the highly-structured source code
text by learning both long-range features and localized
patterns, achieving the best
performance among other competing methods on real-world source code.

\ifdefined\isaccepted
	\section*{Acknowledgements}
	This work was supported by Microsoft Research through
	its PhD Scholarship Programme and
	the Engineering and Physical Sciences
	Research Council [grant number EP/K024043/1].
  We would like to
	thank Daniel Tarlow and Krzysztof Geras for their insightful
	comments and suggestions.
\fi

\bibliographystyle{icml2016}
\bibliography{bibliography}

\end{document}